\documentclass[10pt,twocolumn,letterpaper]{article}

\usepackage{iccv}
\usepackage{times}
\usepackage{epsfig}
\usepackage{graphicx}
\usepackage{amsmath}
\usepackage{amssymb}
\usepackage{booktabs}
\usepackage{color}
\usepackage{bm}
\usepackage{multirow}
\usepackage{flushend}
\usepackage{caption}

\usepackage{algorithm}
\usepackage{algpseudocode}
\usepackage{makecell}
\usepackage{arydshln}
\usepackage{pifont}
\usepackage{booktabs}
\usepackage{mathtools}
\usepackage[breaklinks=true,bookmarks=false]{hyperref}

\iccvfinalcopy % *** Uncomment this line for the final submission

\def\iccvPaperID{} % *** Enter the ICCV Paper ID here

\ificcvfinal\fi

\begin{document}

%%%%%%%%% TITLE
\title{Uncertainty Guided Adaptive Warping for Robust and Efficient Stereo Matching}

\author{
    Junpeng Jing$^{1,2}$\footnotemark[1]\ \ \ Jiankun Li$^{2}$\ \ \  Pengfei Xiong$^{3}$ \ \ \ Jiangyu Liu$^{2}$\ \ \ Shuaicheng Liu$^{2}$\ \ \ 
    \\
    Yichen Guo$^{1}$\ \ \  
    Xin Deng$^{1}$\footnotemark[2]\ \ \ Mai Xu$^{1}$\footnotemark[2]\ \ \ Lai Jiang$^{4}$ \ \ \ Leonid Sigal$^{4}$
    \\
    $^{1}$Beihang University \quad  $^{2}$Megvii Research \quad
    $^{3}$Shopee  \quad $^{4}$University of British Columbia \\
    {\tt\small \{junpengjing, cindydeng, MaiXu\}@buaa.edu.cn }
}

\twocolumn[{
\renewcommand\twocolumn[1][]{#1}
\maketitle\thispagestyle{empty}
    \begin{center}
    % \centering
        % \centering
        \vspace{-.5em}
        \includegraphics[width=1.0\linewidth,page=1]{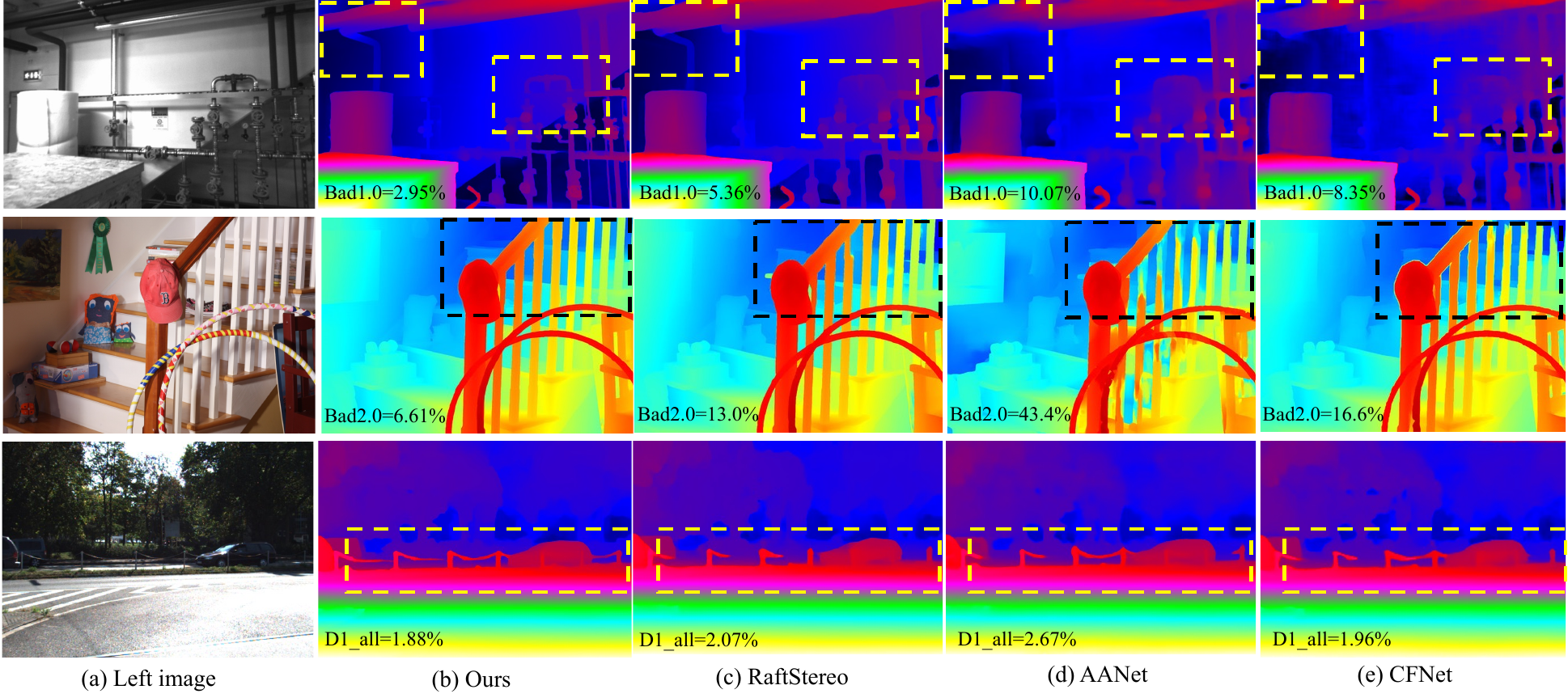}
        \captionof{figure}
        {Visual and quantitative comparisons between our proposed method and other SOTA methods for robust stereo matching on ETH3D, Middlebury, and KITTI 2015 (from top to bottom). All results from each method are directly predicted by a single trained model with fixed parameters without any fine-tuning or adaption. Our method outperforms others both in cross-domain accuracy and details. Obvious errors and bad cases of other methods are highlighted in the box parts.}
        \label{fig:figure1}
    \end{center}
    }
]

% \twocolumn[
% {
%     \maketitle
%     \begin{center}
%     \vspace{-.5em}
%     \includegraphics[width=1.0\linewidth,page=1]{figures/figure1.pdf}
%     \caption{figure}{Visual and quantitative comparisons between our proposed method and other SOTA methods for robust stereo matching on ETH3D, Middlebury, and KITTI 2015 (from top to bottom). All results from each method are directly predicted by a single trained model with fixed parameters without any fine-tuning or adaptation. Our method outperforms others both in cross-domain accuracy and details. Obvious errors and bad cases of other methods are highlighted in the box parts.}
%     \label{fig:figure1}
%     \end{center}
% }
% ]

\maketitle
\ificcvfinal\thispagestyle{empty}\fi

%%%%%%%%% ABSTRACT
\begin{abstract}
Correlation based stereo matching has achieved outstanding performance, which pursues cost volume between two feature maps. Unfortunately, current methods with a fixed model do not work uniformly well across various datasets, greatly limiting their real-world applicability.
\renewcommand{\thefootnote}{\fnsymbol{footnote}}
\newcounter{somecounter}
\setcounter{somecounter}{2}
\footnotetext[1]{Work was done while interning at Megvii. \fnsymbol{somecounter} Corresponding authors.}
\renewcommand{\thefootnote}{\arabic{footnote}}
To tackle this issue, this paper proposes a new perspective to dynamically calculate correlation for robust stereo matching. A novel Uncertainty Guided Adaptive Correlation (UGAC) module is introduced to robustly adapt the same model for different scenarios. Specifically, a variance-based uncertainty estimation is employed to adaptively adjust the sampling area during warping operation. 
Additionally, we improve the traditional non-parametric warping with learnable parameters, such that the position-specific weights can be learned. 
%in the network
We show that by empowering the recurrent network with the UGAC module, stereo matching can be exploited more robustly and effectively. 
Extensive experiments demonstrate that our method achieves state-of-the-art performance over the ETH3D, KITTI, and Middlebury datasets when employing the same fixed model over these datasets without any retraining procedure. To target real-time applications, we further design a lightweight model based on UGAC, which also outperforms other methods over KITTI benchmarks with only 0.6 M parameters.

\end{abstract}

%%%%%%%%% BODY TEXT
\section{Introduction}
\label{sec:intro}
Stereo matching is a fundamental computer vision task \cite{scharstein2002taxonomy} that aims to estimate the disparity between two rectified stereo images. In the past decade, stereo matching has become increasingly popular due to the development of deep learning and the support of large synthetic datasets \cite{mayer2016large, sintel}. As a result, it has a breadth of applications spanning autonomous driving \cite{chen2015deepdriving} to 3D reconstruction \cite{geiger2011stereoscan}. 

Since there are significant domain differences between stereo matching datasets, existing state-of-the-art methods generally fail to achieve robust stereo matching, when applied to different datasets with a single trained model with fixed parameters. As shown in Fig.~\ref{fig:figure1}(a), the Middlebury dataset \cite{middlebury} focuses on indoor scenes with high resolution and large disparity, while ETH3D \cite{eth3d} contains gray-scale images at low resolution, and KITTI \cite{kitti} concentrates on outdoor driving scenarios. Consequently, the leading methods \cite{xu2020aanet,lipson2021raft} on one dataset cannot consistently perform well across different datasets without retraining (Fig.~\ref{fig:figure1}(c,d)), which fails to meet the generalization requirement of real-world applications. 

Large scene differences and unbalanced disparity distribution are the key reasons resulting in noisy and distorted feature maps \cite{zhang2020domain}, thus reducing the robustness. In addition, the limited receptive field of convolutions makes it difficult for the network to capture the global features, leading to domain sensitivity to different datasets \cite{luo2016understanding}. To this end, CFNet \cite{shen2021cfnet} adopted an adaptive disparity range to enlarge the receptive field and alleviate the poor robustness caused by the fixed disparity range. However, it still incurs the issue of robust matching (shown in Fig.~\ref{fig:figure1} (e)) because blurred textures and unclear edges in features still exist when constructing cost volume, which is generated by non-parametric warping and cannot be solved by adjusting the disparity range. Here, features are warped by the corresponding disparities and fixed sampling points in the neighborhood. Because this process utilizes constant weights, it is inherently position-agnostic and cannot capture different feature details, leading to low robustness.

In this paper, we propose an uncertainty guided adaptive correlation module to tackle the above problem, and further develop an advanced cascaded recurrent framework based on CREStereo \cite{li2022practical}, namely CREStereo++, to achieve robust stereo matching. Specifically, towards the problem caused by a fixed sampling area and limited receptive field, we employ a variance-based uncertainty estimation module to adaptively adjust the sampling range in the warping process. Moreover, we improve the traditional non-parametric warping operation with content-adaptive weights. In this way, for those areas with high uncertainty, such as texture-less and occluded parts, the network adopts a wide sampling range. For parts that have achieved accurate matching, a small range of sampling area is suitable enough. Experimentally, as shown in Fig.~\ref{fig:figure1} (b), our method achieves SOTA performances on all three datasets simultaneously without adaptation. To benefit real-time applications, we further propose a lightweight version, namely Lite-CREStereo++, to enable real-time performance. Our Lite-CREStereo++ outperforms all the published real-time methods with less than 60ms inference time on KITTI2012 benchmarks with only 0.6 M parameters.

The main contributions of this paper are as follows:
\begin{itemize}
\item
We introduce a new perspective to calculate correlation dynamically for robust stereo matching that can adapt to various datasets.
\item
We develop an uncertainty guided adaptive warping module that enhances the robustness of the network for different scenarios, which is also valuable in general matching tasks.
\item
We conduct extensive experiments on commonly used benchmarks and achieve SOTA results in terms of both robustness and efficiency, making the proposed approach universal.
\item
Our method obtains the championship on the stereo task of Robust Vision Challenge 2022.
\end{itemize}

%-------------------------------------------------------------------------
\section{Related Works}
\textbf{Deep Stereo Matching.} Recently, the success of convolution neural networks has driven the community to develop learning based solutions for stereo matching \cite{zbontar2015computing, mayer2016large, pang2017cascade, liang2018learning, chang2018pyramid, kendall2017end, guo2019group, lipson2021raft, xu2022attention, li2022practical}. Specifically, Mayer \etal \cite{mayer2016large} proposed the first end-to-end method DispNetC, which directly calculated the correlation between left and right features by multiplying the pixels at the corresponding position. 
Chang \etal introduced PSMNet \cite{chang2018pyramid}, using a spatial pyramid pooling module to leverage the capacity of global context information in different scales.
Based on this, Guo \etal \cite{guo2019group} proposed GwcNet via group-wise correlation, achieving better performance and reducing parameters simultaneously. For most methods, the diversity in disparity distribution is the main challenge for model performance, which can be improved through an iterative mechanism. Following the great success of RAFT \cite{teed2020raft} in optical flow task, RaftStereo \cite{lipson2021raft} was proposed for stereo matching with iterative refinement. Li \etal \cite{li2022practical} proposed CREStereo, which illustrates the effectiveness of cascaded recurrent network. 

\begin{figure*}[t]
   \begin{center}
   %\vspace{-0.5em}
   \includegraphics[width=1\linewidth]{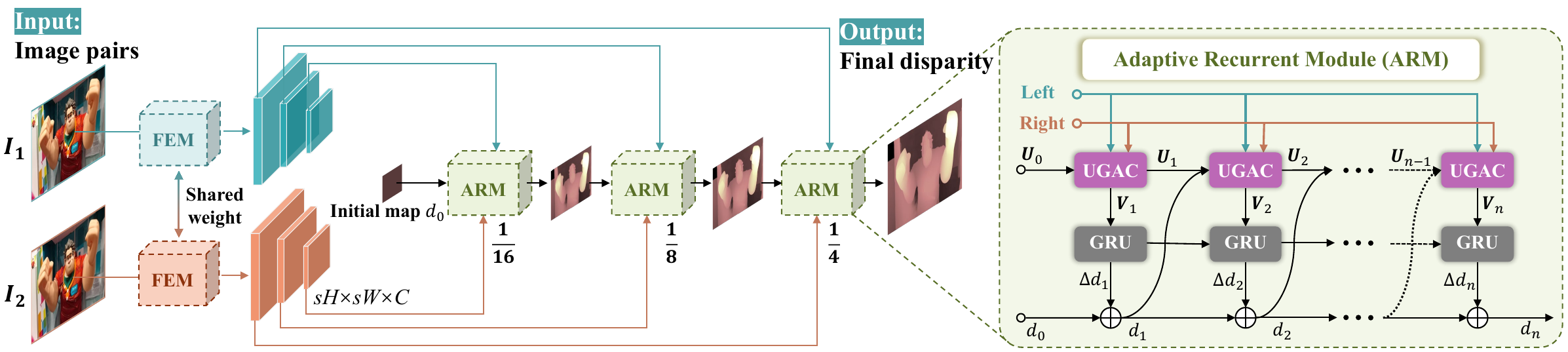}
   \end{center}
   % \vspace{-0.5em}
    \caption{The overall framework of our method. It comprises two shared-weight feature extraction modules (FEM) and a reusable adaptive recurrent module (ARM). Starting from an initial input (a $0$ disparity map), the output of disparity prediction in the former stage is fed to the next ARM. For each iteration in ARM, we first apply Uncertainty Guided Adaptive Correlation (UGAC) to compute the correlation between two features. The disparities are then refined with correlation using Gate Recurrent Unit (GRU). Note that ARM has the same set of parameters and is used repeatedly in each stage.} 
    % \vspace{-1em}
   \label{fig:framework}
\end{figure*}

\textbf{Robust Stereo Matching.} Robust stereo matching oriented toward robustness and real-world applications is a less explored problem. Jia \etal \cite{wang2020improving} introduced an end-to-end network with scene geometry priors to improve the network's generalization ability to unseen scenes. Song \etal \cite{song2021adastereo} introduced a domain adaptation method to handle the gap between synthetic and real-world domains. Zhang \etal \cite{zhang2020domain} proposed a domain-invariant approach via a domain normalization layer and learnable graph-based filter. MCV-MFC \cite{liang2019stereo} proposed a two-stage training strategy to transfer the model to target datasets gently. Shen \etal  \cite{shen2021cfnet} proposed CFNet, a cascaded and fused cost volume based network to deal with the domain difference, illustrating the potential of cascaded architecture for robust vision tasks. However, it still suffers from a lack of flexibility for modeling sampling in complicated structures.

\textbf{Real-time Stereo Matching.} Several recent works \cite{khamis2018stereonet, duggal2019deeppruner, besse2014pmbp, xu2021bilateral, xu2022acvnet} focus on real-time performance while maintaining satisfactory accuracy. StereoNet \cite{khamis2018stereonet} introduced an edge-preserving refinement network to leverage left images to recover high frequency details towards information loss at low resolution. DeepPruner \cite{duggal2019deeppruner} built a sparse cost volume by PatchMatch \cite{besse2014pmbp}, and pruned the search space based on the predicted disparities, which were further refined under the guidance of image features. Xu \etal \cite{xu2020aanet} proposed AANet, designed a sparse points based cost aggregation method and replaced the commonly used 3D convolutions to achieve fast inference speed. Xu \etal \cite{xu2022acvnet} introduced Fast-ACVNet, which adopted an attention mechanism to suppress redundant information and enhance matching-related information in the concatenation volume, which is quite efficient. In this paper, we also introduce a real-time stereo matching network based on the proposed network architecture while maintaining accuracy.

%------------------------------------------------------------------------
\section{Methods}

\subsection{Overall Framework}
Inspired by \cite{shen2021cfnet}, a cascaded network is developed in our method to predict disparity from a low resolution to a high resolution. This way, a larger receptive field can be obtained to better learn the global structural representations. Precisely, as shown in Fig.~\ref{fig:framework}, we follow the framework in \cite{li2022practical} and design a much simplified cascaded backbone, which is composed of only two basic components without any aggregation or attention mechanism.

Given an input image pair of $\mathbf{I}_{L}$ and $\mathbf{I}_{R}$ where $\mathbf{I}_{L}, \mathbf{I}_{R} \in \mathbb{R}^{H\times W\times 3}$, two share-weighted feature extraction modules (FEM) are employed to pyramidally  extract multi-scale features $\left \{\mathbf{F}_{L}^s \right \} , \left \{\mathbf{F}_{R}^s \right \} \in \mathbb{R}^{sH\times sW\times C}$. Note that $s \in \left \{1/4, 1/8, 1/16 \right \}$ represents the set of down-sampled scales and $C$ is the channel number. Then, the multi-scale features pass through 3 cascaded stages of the proposed adaptive recurrent module (ARM), which is composed of an uncertainty guided adaptive correlation (UGAC) module and a gate recurrent unit (GRU) \cite{cho2014properties}. In the ARM, the cost volume is calculated via UGAC and then input into GRU, for iteratively refining the disparity prediction results. To simultaneously enhance the robustness and preserve the details of the input, the final disparity prediction of ARM in each stage is adopted as the initial disparity of the GRU in the next stage. Note that the ARMs used in 3 cascaded stages share the same parameters, which shows a high potential to implement a lightweight model. Finally, the predicted disparity at the last stage is up-sampled to the original resolution by convex up-sampling \cite{teed2020raft}.

\begin{figure*}[t]
   \begin{center}
   % \vspace{-0.5em}
   \includegraphics[width=1\linewidth]{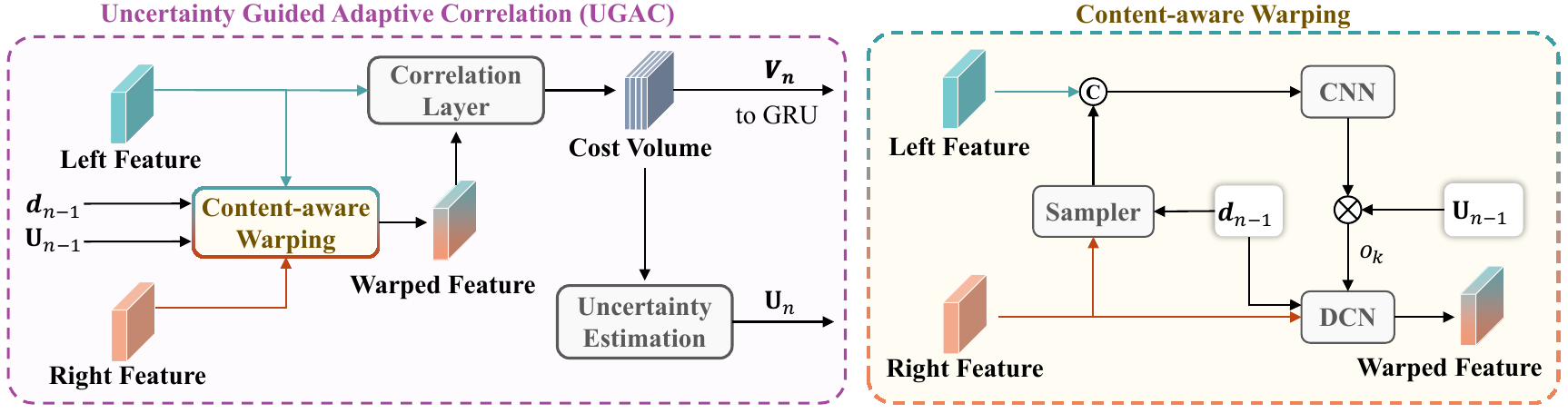}
   \end{center}
   % \vspace{-1.5em}
    \caption{Left: The architecture of our uncertainty guided adaptive correlation (UGAC), is composed of a content-aware warping layer, correlation layer, and uncertainty estimation. Right: The workflow of the content-aware warping layer, where DCN represents deformable convolution network, CNN is three 3$\times$3 convolutions with leakyReLU following each layer. For simplicity, we take the UGAC of the $n$-th iteration as the example.}
   \label{fig:warping}
   % \vspace{-1em}
\end{figure*}
\subsection{Uncertainty Guided Adaptive Correlation}

As shown in Fig.~\ref{fig:warping}, the UGAC module consists of a content-aware warping layer, a correlation layer, and uncertainty estimation. For the $n$-th iteration of ARM, the right features $\left \{\mathbf{F}_{R}^s \right \}$ are first warped via the content-aware warping layer, considering the prediction disparity $\bm{d}_{n-1}$ and the uncertainty map $\bm{U}_{n-1}$ at the $(n-1)$-th iteration. Then, the cost volume $\bm{V}_{n}$ between the left features and the warped right features is calculated by the correlation layer. Given the cost volume $\bm{V}_{n}$, the uncertainty map $\bm{U}_{n}$ is estimated and fed to the UGAC of the next iteration.  Note that $\bm{V}_{n}$ is used as the input of GRU. 

\textbf{Correlation Layer.} In the correlation layer, the cost volume is calculated on the top of local correlation mechanism. Specifically, the cost volume $\bm{V}_n$ at position $\bm{p}$ can be formulated as follows,
\begin{equation}
\bm{V}_n(\bm{p}) =\sum_{\bm{r} \in \bm{R}} \left \langle \mathbf{F}_{L}(\bm{p}) \cdot \mathbf{F}_{R}(\bm{p} + \bm{r}) \right \rangle, 
\end{equation} 

where $\bm{R}$ denotes the search range of the current pixel in specific directions, and $\left \langle \cdot \right \rangle $ represents the channel-wise product operation.

\textbf{Content-aware Warping Layer.} 
In existing methods based on PWCNet \cite{sun2018pwc}, the warping layer warps right features $\mathbf{F}_{R}$ towards the left features $\mathbf{F}_{L}$ via current disparity $\bm{d}_{n-1}$ to obtain the warped right features $\mathbf{\hat{F}}_{R}$, formulated as follows:
\begin{equation}
\mathbf{\hat{F}}_{R}(\bm{p})=\sum_{\bm{k} \in \bm{K}} \bm{c}_k \cdot \mathbf{F}_{R}\left(\bm{p}+ \bm{d}_{n-1}({\bm{p} + \bm{k}})\right),
\label{eq:traditional warping}
\end{equation} 
where $\bm{K}$ denotes the sampling point area centered on pixel $\bm{p}$, and $\bm{d}_{n-1}({\bm{p} + \bm{k}})$ represents the corresponding disparity at the position of $\bm{p} + \bm{k}$. Besides, $\bm{c}_k$ is the weight for the $k$-th point, usually set as a constant.

However, the above equation neglects the diversity in the warping process and adopts a content-agnostic treatment for all cases. It is thereby tough to implement ``perfect" warping, leading to distorted and noisy features due to the mismatching caused by occlusions, non-texture and repetitive-texture areas. We thus calculate $\bm{c}_k$ in a content-specific manner, denoted as $\bm{w}_k\left(\bm{p}\right)$. Moreover, considering the different disparity ranges and distributions in different cases, it is reasonable to adopt different sampling ranges to alleviate the domain-sensitive problem. To this end, we introduce an extra offset $\bm{o}({\bm{p,k}})$ to expand the sampling range and achieve learnable warping formulated as follows:
\begin{equation}
\mathbf{\hat{F}}_{R}(\bm{p})=\sum_{\bm{k} \in \bm{K}} \bm{w}_k\left(\bm{p}\right) \cdot \mathbf{F}_{R}\left(\bm{p}+ \bm{d}_{n-1}({\bm{p}+\bm{k}}) + \bm{o}({\bm{p,k}})\right),
\label{eq:adaptive warping}
\end{equation} 
which is achieved with group-wise deformable convolutions \cite{dai2017deformable} in practice.

\textbf{Uncertainty Estimation.} The ambiguity caused by traditional non-parametric warping usually accounts for a small proportion of each sample. Therefore, we expand the sampling range of ill-posed pixels and conduct adaptive pixel-level adjustments. Previous works \cite{shen2021cfnet, kendall2017end} have observed that ill-posed areas, texture-less regions, and occlusions tend to be multi-modal distributions with a high estimation error rate. Motivated by this, we introduce a variance-based uncertainty estimation to guide the offset $\bm{o}({\bm{p,k}})$, and further balance the disparity distributions of different datasets, which is formulated as follows,
\begin{equation}
\bm{U}_n =1 - \sigma ( \sum\left(\bar{\bm{V}}_n-\bm{V}_n\right)^{2}), \\
\end{equation} 
\begin{equation}
\bm{o} = \bm{U}_n \cdot \text{CNN}[\mathbf{F}_{L}, \mathcal{S}(\mathbf{F}_{R}, \bm{d}_{n-1})],
\end{equation} 
where $\bm{V}_{n}$ is cost volume and $\bm{S}$ represents bilinear sampler, $\bar{\bm{V}}_n$ represents the average value of $\bm{V}_n$, $\sigma(\cdot)$ is the sigmoid function. Through this, the network can leverage the prior knowledge of disparity prediction at the current iteration to adaptively capture more possible sampling objects.

\begin{table*}[htbp]
  \centering
  \small
    \caption{Ablation study of the proposed method (Lite-CREStereo++) on Middlebury, KITTI2015, and ETH3D dataset. The  network component  is evaluated individually in each section of the table and the approach used in our final model is underlined. UE: uncertainty estimation. Inference time is measured on KITTI by V100 GPU.}
    % \vspace{-1em}
  \setlength{\tabcolsep}{5.pt}
  \begin{tabular}{ccccccccc}
    \toprule
    \multirow{2}{*}[-2pt]{Experiment} &\multirow{2}{*}[-2pt]{Method} & \multicolumn{2}{c}{Middlebury} & \multicolumn{2}{c}{ETH3D} & \multicolumn{1}{c}{KITTI15} & \multicolumn{1}{c}{\multirow{2}{*}{Params.(M)}} & \multicolumn{1}{c}{\multirow{2}{*}{Runtime(ms)}}\\ 
    \addlinespace[-12pt] \\
    \cmidrule(lr){3-7}
    \addlinespace[-12pt] \\ 
    & & Bad 2.0 & AvgErr & Bad 1.0 & AvgErr & D1-all & \\
    \midrule
    \multirow{4}{*}[-2pt]{GRU Kernel Size} & $3\times3$ & 14.21 & 2.78 & 2.30 & 0.27 & 2.43 & \textbf{0.514} & \textbf{47.9}\\
    & $1\times5$ & 12.95 & 2.61 & 2.34 & 0.27 & 2.42 & 0.523 & 53.1 \\
    & $1\times15$ & 12.02 & \textbf{2.46} & 2.10 & \textbf{0.25}  & 2.38 & 0.584 & 54.9 \\
    & \underline{$1\times5$ + $1\times15$} & \textbf{11.86} & \textbf{2.46} & \textbf{1.98} & 0.26 &  \textbf{2.36} & 0.595 & 56.2 \\
    & $1\times5$ + $1\times15$ + $1\times31$ & 11.98 & 2.55 & 2.00 & 0.26  &  2.39 & 0.745 & 63.3 \\
    \midrule
    \multirow{3}{*}[-2pt]{Warping} & Bilinear & 14.79 & 2.80 & 2.31 & 0.27 & 2.55  &  \textbf{0.527} & \textbf{41.0}\\
    & Content-aware & 12.96 & \textbf{2.46} & 2.02 & \textbf{0.26} & 2.47  & 0.595 & 55.8\\
    & \underline{UE + Content-aware} & \textbf{11.86} & \textbf{2.46} & \textbf{1.98} & \textbf{0.26} & \textbf{2.36}  & 0.595 & 56.2\\
    \midrule
    \multirow{3}{*}[-2pt]{Uncertainty Estimation} &Variance + Tanh & 11.98 & 2.46 & 2.01 & \textbf{0.25} & 2.42  & 0.595 & \textbf{56.2} \\
    & \underline{Variance + Sigmoid} & \textbf{11.86} & \textbf{2.46} & \textbf{1.98} & 0.26 & \textbf{2.36} & \textbf{0.595} & \textbf{56.2}\\
     & Error-aware + Sigmoid & 12.40 & 2.49 & 2.05 & \textbf{0.25} & 2.40  & 0.595 & 57.1 \\
    \bottomrule
  \end{tabular}
  \label{tab:ablation}
  % \vspace{-0.5em}
\end{table*}

\textbf{Comparison with Existing Adaptive Mechanisms.} It is worth noting that other works \cite{xu2020aanet,li2022practical} also leverage the idea of adaptive mechanisms. Here, we emphasize the critical difference in our method. In previous typical
methods, some works \cite{stereo1994, robust2011} calculate adaptive weights for correlation or adaptively control the window size in correlation. In AANet \cite{xu2020aanet}, the adaptive aggregation is conducted after warping, where a set of deformable convolutions are developed to replace the original convolutions. However, the cost volume is still built via traditional warping operation in Eq.~\ref{eq:traditional warping}, which 
still embeds the error during the alignment of two features. Therefore, it is necessary to refine the features in the warping process.
% Even if they are refined in the subsequent process, the performance is not as effective as that of direct intervention in the warping process.
In CREStereo \cite{li2022practical}, the cost volume is calculated with an adaptive position based on the local correlation. The adaptive mechanism is applied to change the matching window shape. It is ineffective to adapt the position only on the warped features. 
Compared with these approaches, our method conducts effective adaptation during warping before building the cost volume, alleviating the blur and inaccurate problems caused by occlusions and texture-less areas from the source. We visualize the difference between the traditional warping and ours, as shown in Fig.~\ref{fig:visual-warping}. In addition, we make full use of the prior information to produce an uncertainty map, that the adaptive mechanism is guided by a variance-based uncertainty estimation instead of directly learned by convolutions, which makes the adaptive process more reasonable and stable.

\subsection{Lite-CREStereo++} We also design a lightweight version of the proposed model, namely Lite-CREStereo++, which adopts the same backbone but with fewer channels and iteration numbers. Specifically, the channel number $C$ is reduced to $64$ from $256$ in the feature extraction module, which is also reduced correspondingly in the following ARM. To achieve real-time disparity prediction without sacrificing too much accuracy, we introduce an extra convolution layer with a super kernel size $1 \times 15$ in GRU, which improves the accuracy with little extra cost. The effectiveness of the lightweight model is verified in Sec.~\ref{Sec:Ablation Study}. Besides, different from the slow-fast setting in RaftStereo \cite{lipson2021raft}, we increase the iteration numbers of ARM from small resolution to large resolution instead. In detail, the iteration numbers are set as $2,4$, and $6$, respectively. In this way, we achieve a competent balance between accuracy and speed.

\subsection{Loss Function}
We supervise the optimization with $l_1$ distance between the ground truth disparities and the predictions to train the model in an end-to-end manner. All disparity predictions in all GRU cells are supervised with ground truth in training, while only the last disparity prediction is obtained as the final output. The total loss is formulated as follows:
\begin{equation}
\mathcal{L} = \sum_{s} \sum_{i=1}^{n} 
\gamma^{n - i} || \mathbf{d}_{\mathrm{gt}} - \mathcal{S}(\mathbf{d}_{i}^{s}) ||_1,
\end{equation}
where the exponentially weight $\gamma$ is set to 0.8, and $\mathcal{S}(\mathbf{d}_{i}^{s})$ represents the predictions after sampler $\mathcal{S}$.

% \begin{figure}[htbp]
%   \begin{center}
%   \includegraphics[width=1\linewidth]{figures/visual-warping.pdf}
%   \end{center}
%       \caption{Illustration of the difference between existing methods and our proposed method.} 
%   \label{fig:xx}
% \end{figure}

% \begin{table}[t]
%   \centering
%   \small
%     \caption{
%   Ablation study for different input size during inference.}
%   \setlength{\tabcolsep}{3.pt}
%   \begin{tabular}{lccccc}
%     \toprule
%     \multirow{2}{*}[-2pt]{Method} & \multicolumn{2}{c}{Middlebury} & \multicolumn{2}{c}{ETH3D}\\ 
%     \addlinespace[-12pt] \\
%     \cmidrule(lr){2-3} \cmidrule(lr){4-5} 
%     \addlinespace[-12pt] \\ 
%     & Bad 2.0 & AvgErr & Bad 1.0 & Avgerr\\
%     \midrule
    
%     $384 \times 512$ &   &   &   &    \\
%     $768 \times 1024$ &   &   &   &   \\
%     $1536 \times 2048$ &   &   &   &  \\

%     \bottomrule
%   \end{tabular}
%   \label{tab:ablation2}

% \end{table}

\begin{table*}[t] \footnotesize\addtolength{\tabcolsep}{-2pt}
\centering
     \caption{Robustness comparison among ETH3D, Middlebury, and KITTI2015 testsets with existing SOTA methods in RVC. All methods are tested on three datasets with a single fixed model. The overall rank is obtained by Schulze Proportional Ranking \cite{schulze2011new} to combine multiple rankings into one. Our approach achieves the best overall performance.}
    % \vspace{-1em}
    \begin{tabular}{c|cccc|cccc|cccc|c} 
    \toprule
    \multirow{2}{*}{Method} & \multicolumn{4}{c|}{Middlebury} & \multicolumn{4}{c|}{KITTI2015} & \multicolumn{4}{c|}{ETH3D} & {Overall}  \\ 
    \cline{2-13}
                            & bad 1.0 & bad 2.0 & AvgErr & Rank         & D1-bg & D1-fg & D1-all & Rank        & bad 1.0 & bad 2.0 & AvgErr & Rank   & Rank     \\ 
    \hline
    AANet\_RVC \cite{xu2020aanet}                       & 42.9 & 31.8 & 12.8 & 10            & 2.23 & 4.89 & 2.67 & 10         & 5.41 & 1.95 & 0.33 & 10      & 12   \\
    
    CVANet\_RVC                                         & 58.5 & 38.5 & 8.64 & 11            & 1.74 & 4.98 & 2.28 & 9         & 4.68 & 1.37 & 0.34 & 9      & 11   \\ 
    
    GANet\_RVC \cite{zhang2019ga}                       & 43.1 & 24.9 & 15.8 & 11            & 1.88 & 4.58 & 2.33 & 7         & 6.97 & 1.25 & 0.45 & 10      & 10   \\
    
    HSMNet\_RVC \cite{yang2019hierarchical}             & 31.2 & 16.5 & 3.44 & 6            & 2.74 & 8.73 & 3.74 & 12         & 4.40 & 1.51 & 0.28 & 8      & 9   \\ 
    
    MaskLacGwcNet\_RVC \cite{guo2019group}              & 31.3 & 15.8 & 13.5 & 8            & 1.65 & 3.68 & 1.99 & 5         & 6.42 & 1.88 & 0.38 & 12       & 8     \\
    
    GEStereo\_RVC                                       & 22.8 & 14.1 & 3.78 & 3            & 2.29 & 4.79 & 2.71 & 11         & 3.95 & 1.25 & 0.29 & 6       & 7    \\
    
    CroCo\_RVC                                          & 32.9 & 19.7 & 5.14 & 9            & 2.04 & 3.75 & 2.33 & 7         & \textbf{1.54} & \underline{0.50} & 0.21 & \underline{2}       & 6    \\
    
    NLCANet\_V2\_RVC \cite{rao2020nlca}                 & 29.4 & 16.4 & 5.60 & 7            & \textbf{1.51} & 3.97 & 1.92 & 3         & 4.11 & 1.20 & 0.29 & 6      & 5    \\ 
    
    CFNet\_RVC \cite{shen2021cfnet}                     & 26.2 & 16.1 & 5.07 & 5            & 1.65 & 3.53 & 1.96 & 3         & 3.70 & 0.97 & 0.26 & 5      & 4    \\
    
    iRaftStereo \_RVC \cite{lipson2021raft}             & 24.0 & \underline{13.3} & \underline{2.90} & \underline{2}            & 1.88 & \underline{3.03} & 2.07 & 6         & 1.88 & 0.55 & \underline{0.17} & 3       & 3   \\
    
    raft+\_RVC \cite{teed2020raft}                 & \underline{22.6} & 14.4 & 3.86 & 4            & 1.60 & \textbf{2.98} & \textbf{1.83} & \textbf{1}         & 2.18 & 0.71 & 0.21 & 4       & \underline{2}    \\
    
    CREStereo++\_RVC (ours)                             & \textbf{16.5} & \textbf{9.46} & \textbf{2.20}  & \textbf{1}            & \underline{1.55} & 3.53 & \underline{1.88} & \underline{2}         & \underline{1.70} & \textbf{0.37} & \textbf{0.16} & \textbf{1}       & \textbf{1}     \\
    
\bottomrule
\end{tabular}
\label{tab:comparison_rvc}
% \vspace{-1em}
\end{table*}

\section{Experiments}
More details about datasets, implementation, and evaluation can be seen in the supplementary materials.

\subsection{Datasets}

For training, several public datasets are used, including Middlebury \cite{scharstein2014high}, ETH3D
\cite{eth3d}, KITTI \cite{menze2015object}, SceneFlow \cite{mayer2016large}, Sintel \cite{sintel}, Falling Things \cite{fallingthings}, InStereo2K
\cite{bao2020instereo2k}, Carla \cite{deschaud2021kitti}, and the dataset proposed in \cite{li2022practical}. For evaluation, following the previous methods, we adopt the commonly used benchmarks, including Middlebury 2014 \cite{scharstein2014high} (full resolution), ETH3D
\cite{eth3d}, and KITTI 2012/2015 \cite{menze2015object}.

\subsection{Implementation Details}

\begin{figure}[t]
   \begin{center}
   \includegraphics[width=0.83\linewidth]{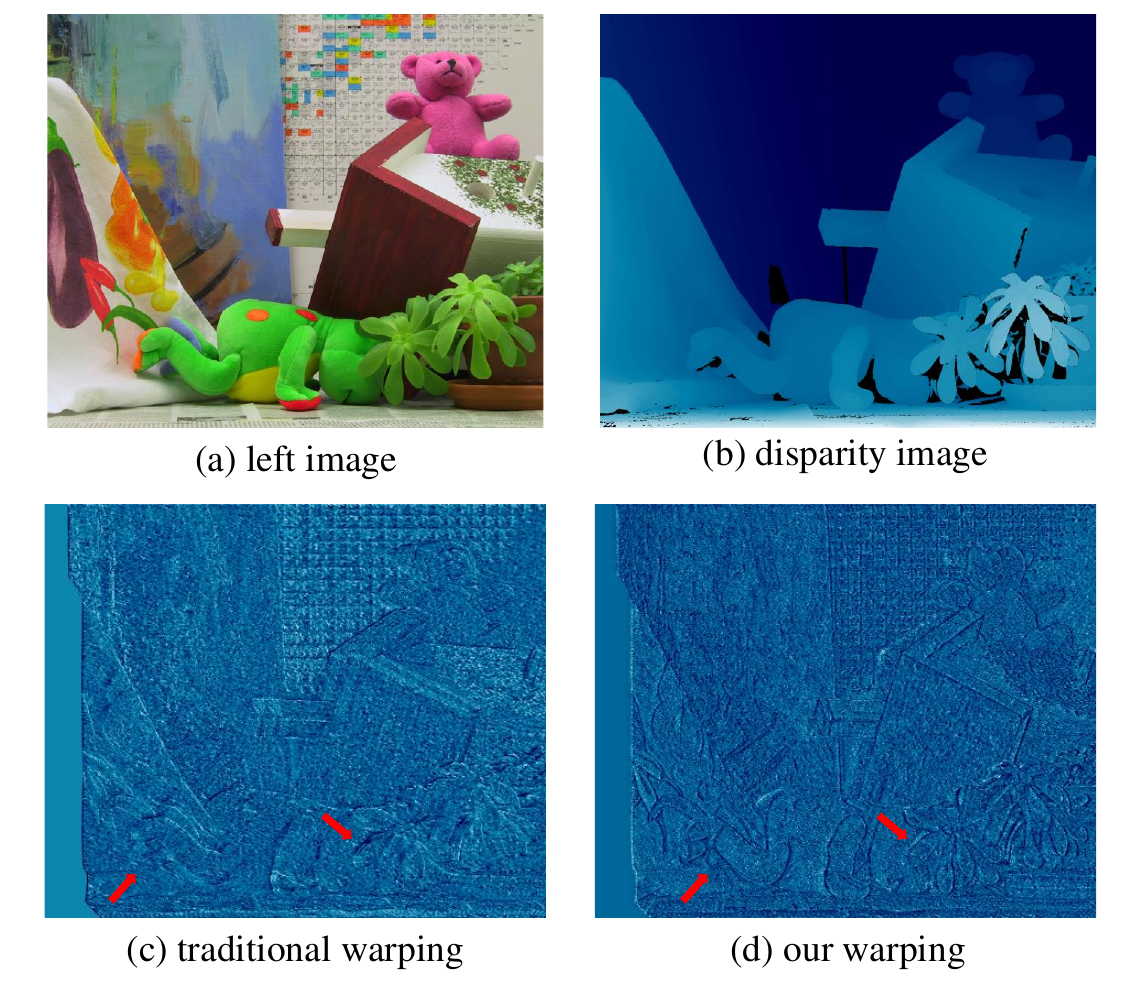}
   \end{center}
    % \vspace{-.8em}
    \caption{Visual comparison between traditional warping and ours. The feature warped by traditional method has obviously blurry edges and distortions, while warped by our method has sharper details, as indicated by the red arrows.} 
    \label{fig:visual-warping}
    % \vspace{-0.9em}
\end{figure}

We conduct a two-stage training strategy to train the proposed method. First, during the pre-training process, all the datasets above are used except KITTI. Since the ground truth of KITTI is sparse, with more than $1/4$ pixels masked, adding KITTI at an early stage will reduce the overall strength of supervised training, thereby weakening the network's ability of precise matching. Thus, we remove KITTI from the trainset at the pre-training stage. Due to the different quantities of each dataset, we have balanced their proportions in advance. The number of iterations is set to 150k with a learning rate of $4 \times 10^{-4}$ using Adam \cite{adam} optimizer with $\beta_{1}=0.9$ and $\beta_{2}=0.999$. To balance input data from various aspect ratios, the group of stereo images and disparity are first resized to a similar size and then cropped to $384 \times 512$. Second, in the fine-tuning stage, KITTI 2012/2015 is reintroduced for another 50k iterations with a much lower learning rate $1 \times 10^{-4}$ after model convergence, and its proportion is adjusted to half of the whole training set. Considering the aspect ratio in KITTI is much larger than other datasets ($>$ 3), during the fine-tuning stage, the input size is set to $256 \times 512$ for CREStereo++\_RVC and $384 \times 1248$ for Lite-CREStereo++.

\subsection{Ablation Study \label{Sec:Ablation Study}}
As shown in Table~\ref{tab:ablation}, we study a specific component of our approach in isolation and underline the settings used in final model. Experiments are conducted on the lite model.
% All experiments are conducted on the lite model, including all the datasets. 

\textbf{GRU Kernel Size.} We explore the effect of different kernel sizes in GRU. Specifically, the kernel size is increased from $1 \times 5$ to $1 \times 31$, growing $2\times$ each time with different combination ways. The commonly used $3 \times 3$ kernel is also tested. From the table ``GRU Kernel Size", we can see the combination of  $1 \times 5$ and $1 \times 15$ achieves the best overall performance. Although it takes 8ms longer time consumption than $3 \times 3$ kernel, it achieves 2.35 and 0.32 improvement on Middlebury and ETH3D, respectively.

\textbf{Warping Types.} In order to compare the performance of different types of warping, we replace our warping layers with other forms. Specifically, ``UE" represents uncertainty estimation and ``Content-aware" denotes the warping operation in Eq.~\ref{eq:adaptive warping}. As shown in the table, the proposed uncertainty guided adaptive warping achieves the best performance with an acceptable computation complexity. Besides, compared with the traditional bilinear warping operation, learnable warping without uncertainty still has significant advantages, which illustrates the effectiveness of the proposed uncertainty estimation and deformable warping. Visualization of the difference between traditional warping and ours can be seen in Fig.~\ref{fig:visual-warping}. Compared to traditional warping, our
method has obviously sharper feature details. The original method causes the warping of hair and leg areas to be misled by the background.

\textbf{Uncertainty Estimation.} We also explore the effect of different uncertainty estimation approaches, as shown in the table ``Uncertainty Estimation". Error map means the guided map for deformable warping is calculated directly by the difference between left image and warped right image, which has limited improvements. Variance-based approach has better performance. It can also be observed that variance with sigmoid is slightly better than with tanh.

\begin{table}[t]
  \centering
  \small
    \caption{Cross-domain robustness evaluation on ETH3D, Middlebury, and KITTI2012/2015 trainsets. All methods are only trained on the Scene Flow dataset and evaluated on each dataset with fixed parameters.}
  % \vspace{-1em}
  \setlength{\tabcolsep}{1.6pt}
  \begin{tabular}{l|ccccc}
    \toprule
    \multirow{2}{*}[-2pt]{Method} & {Middlebury} & {KITTI2012} & {KITTI2015}& {ETH3D}\\ 
    \addlinespace[-12pt] \\
    % \cline{2-5}
    \addlinespace[-12pt] \\ 
    & bad 2.0  &  D1-all &  D1-all & bad 1.0 \\
    \midrule
    PSMNet \cite{chang2018pyramid} & 39.5 & 15.1 & 16.3 & 23.8 \\
    GWCNet \cite{guo2019group} & 37.4 & 12.0 & 12.2 & 11.0  \\
    CasStereo \cite{gu2020cascade} & 40.6 & 11.8 & 11.9 & 7.8  \\
    GANet \cite{zhang2019ga} & 32.2 & 10.1 & 11.7 & 14.1 \\
    DSMNet \cite{zhang2020domain}& 21.8 & \underline{6.2} & 6.5 & 6.2 \\
    LEAStereo \cite{cheng2020hierarchical} & 31.3 & 9.0 & 9.4 & 9.0 \\
    CFNet \cite{shen2021cfnet} & 28.2 & \textbf{4.7} & 5.8 & 5.8 \\
    RAFT-Stereo \cite{lipson2021raft} & 21.6 & \textbf{4.7} & \underline{5.5} & 7.8 \\
    CREStereo \cite{li2022practical}& \underline{15.3}  & 6.7 & 6.7 & \underline{5.5} \\
    CREStereo++(ours) & \textbf{14.8} & \textbf{4.7} & \textbf{5.2} & \textbf{4.4} \\

    \bottomrule
  \end{tabular}
  \label{tab:crossdomain}
  % \vspace{-1em}
\end{table}

\begin{table*}[t] \footnotesize\addtolength{\tabcolsep}{-2pt}
\centering
\caption{Quantitative evaluation of real-time stereo matching on the online test sets of KITTI 2012 and KITTI 2015. We adopt other SOTA real-time approaches to illustrate the efficiency of the proposed Lite-CREStereo++.}
% \vspace{-1em}
\begin{tabular}{c|cccccc|ccc|c|c} 
\toprule
\multirow{2}{*}{Method} & \multicolumn{6}{c|}{KITTI 2012} & \multicolumn{3}{c|}{KITTI 2015} & \multicolumn{1}{|c}{\multirow{2}{*}{Params.(M)}} & \multicolumn{1}{|c}{\multirow{2}{*}{Runtime(ms)}}  \\ 
\cline{2-10}
& 3-noc & 3-all & 4-noc & 4-all & EPE-noc & EPE-all        & D1-bg & D1-fg & D1-all  &   &  \\ 
\hline
DispNetC \cite{mayer2016large}            & 4.11 & 4.65 & 2.77 & 3.20 & 0.9 & 1.0  & 4.32 & 4.41 & 4.34      & 42.32 &  60    \\

DeepPrunerFast \cite{duggal2019deeppruner}& -- & -- & -- & -- & -- & -- & 2.32 & 3.91 & 2.59      & 7.39 & 50 \\ 

AANet \cite{xu2020aanet}                  & 1.91 & 2.42 & 1.46 & 1.87 & 0.5 & 0.6  & 1.99 & 5.39 & 2.55      & 3.93 & 60 \\

DecNet \cite{yao2021decomposition}        & -- & -- & -- & -- & -- & -- &  2.07 & 3.87 & 2.37                & -- & 50 \\

BGNet \cite{xu2021bilateral}              & 1.77 & 2.15 & --  & -- & 0.6 & 0.6     & 2.07 & 4.74 & 2.51      & 2.97 & {44} \\

BGNet+ \cite{xu2021bilateral}             & 1.62 & 2.03 & 1.16 & 1.48 & 0.5 & 0.6  & 1.81 & 4.09 & 2.19      & 5.31 & 48 \\

CoEx \cite{bangunharcana2021correlate}    & 1.55 & 1.93 & 1.15 & \textbf{1.42} & 0.5 & 0.5  & \underline{1.79} & 3.82 & 2.13      & 2.70 & \underline{33} \\

HITNet \cite{tankovich2021hitnet}         & \textbf{1.41} & \underline{1.89} & \underline{1.14} & 1.53 & 0.4 & 0.5  & \textbf{1.74} & \textbf{3.20} & \textbf{1.98}      & \underline{0.63} & \textbf{31} \\

Fast-ACVNet \cite{xu2022acvnet}           & 1.68 & 2.13 & 1.23 & 1.56 & 0.5 & 0.6  & 1.82 & 3.93 & 2.17      & 3.08 & 45 \\

Lite-CREStereo++ (ours)                   & \underline{1.43} & \textbf{1.82} & \textbf{1.12} & \underline{1.44} & 0.5 & 0.5  & \underline{1.79} & \underline{3.53} & \underline{2.08}      & \textbf{0.60} & 56 \\

\bottomrule
\end{tabular}
  \label{tab:Efficiency evaluation}
% \vspace{-0.8em}
\end{table*}

\subsection{Robustness Evaluation}
Robustness measures the generalization ability of a model using a specific set of parameters, which is of great significance in practical applications. Many existing methods are limited to a specific area and only make steady progress on each individual dataset, but cannot obtain comparable results on multiple datasets. To this end, we conduct robustness experiments. More experimental results can be seen in supplementary materials.

\textbf{Domain Transfer Evaluation.} Table.~\ref{tab:comparison_rvc} displays the results of our method and existing SOTA methods in stereo matching of robust vision challenge (RVC). We conduct comparison experiments following previous RVC settings in CFNet\cite{shen2021cfnet}. In RVC, all methods are evaluated on three real-world public benchmarks with a single fixed model, that has the same model parameters without fine-tuning. As can be seen, raft+\_RVC achieves 1st on KITTI2015 among all the methods. However, it fails to obtain comparable results on the other two datasets (4th on ETH3D and Middlebury, respectively), which are far worse than the other top three methods. Similar situations occur in iRaftStereo\_RVC, which ranks 2nd on Middlebury and 3rd ETH3D, but ranks 6th on KITTI2015. In contrast, our method shows strong robustness ability and performs well on all three datasets. We get 1st place on ETH3D and Middlebury, outperforming other methods with a large margin, and 2nd on KITTI2015, achieving the best overall performance. Visual comparisons are shown in Fig.~\ref{fig:figure1}. In the ETH3D samples, it can be seen that other methods have noticeable disparity distortion at the location of the water pipe (yellow box in the figure). In comparison, our method produces sharper object boundaries and better preserves the overall structures. Similar phenomena exist in Middlebury and KITTI2015.

%\vspace{-0.5em}
\begin{figure}[t]
  \begin{center}
  \includegraphics[width=.88\linewidth]{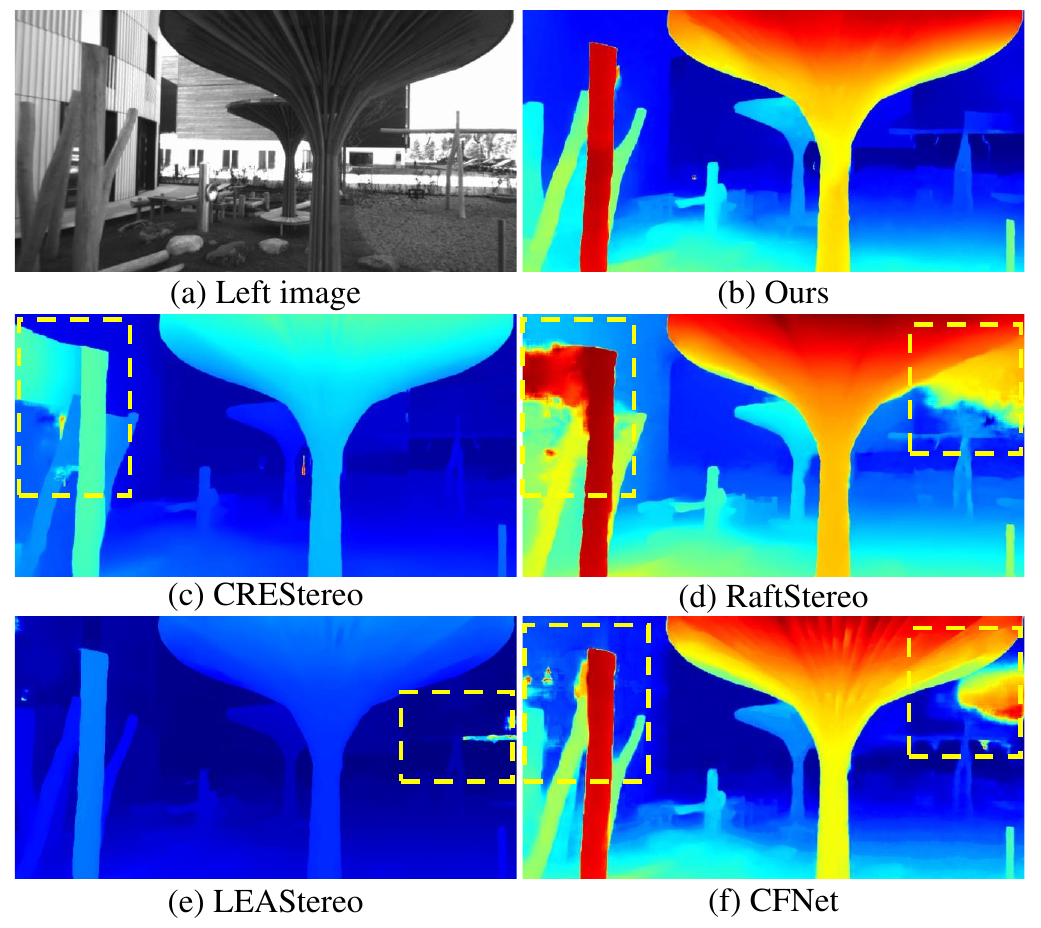}
  \end{center}
% \vspace{-1.7em}
      \caption{Visual comparisons on ETH3D train sets with existing SOTA methods. All models are trained only on Scene Flow. Zoom in for a best view.} 
  \label{fig:visual-comparision}
% \vspace{-1em}
\end{figure}

% \vspace{-0.5em}
\begin{figure}[t]
  \begin{center}
  \includegraphics[width=.88\linewidth]{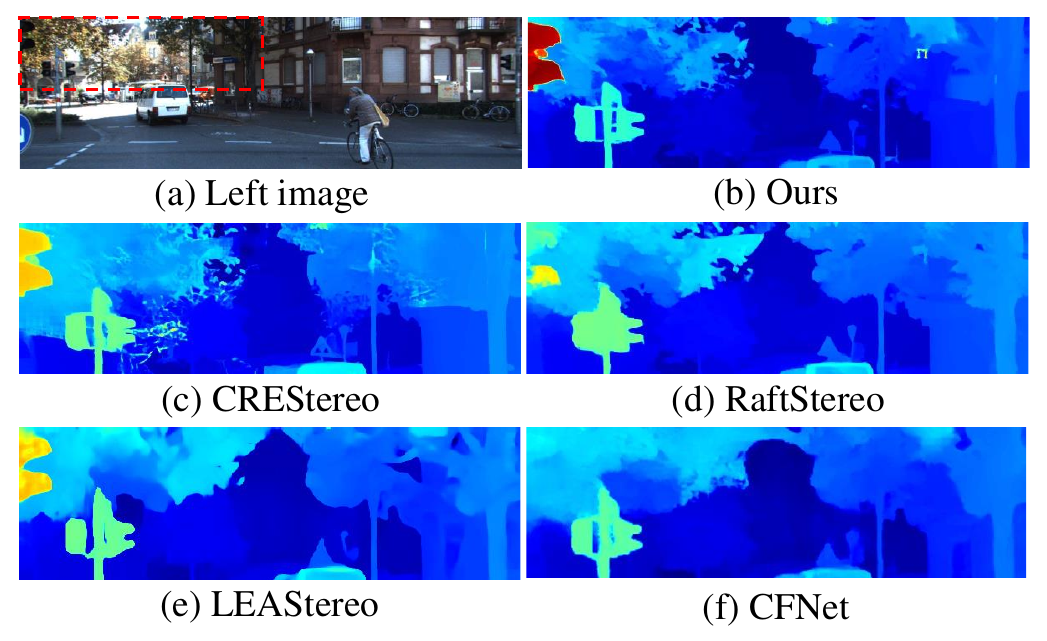}
  \end{center}
     % \vspace{-1.7em}
      \caption{Visual comparisons on KITTI2015 train sets with existing SOTA methods. All models are trained only on Scene Flow. Zoom in for a best view.} 
  \label{fig:visual-comparision2}
% \vspace{-1em}
\end{figure}

\begin{figure*}[t]
   \begin{center}
   \includegraphics[width=.86\linewidth]{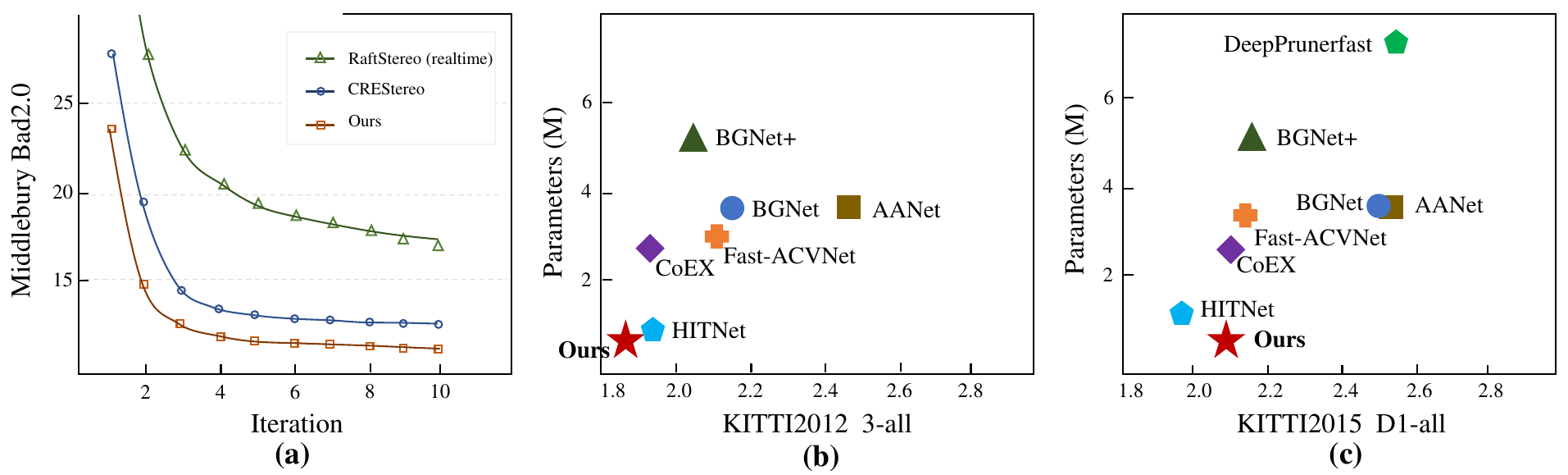}
   \end{center}
   \vspace{-.8em}
      \caption{(a) Iterations vs. Bad 2.0 on the Middlebury dataset. (b) 3-all errors vs. Parameters on the KITTI 2012 leaderboard. (c) D1-all error vs. Parameters on the KITTI 2015 leaderboard. Our method outperforms other SOTA methods.} 
   \label{fig:efficiency}
% \vspace{-0.7em}
\end{figure*}

\textbf{Cross Domain Evaluation.} Following the experiments in \cite{shen2021cfnet}, we conduct cross-domain generalization evaluation to further emphasize the effectiveness of our method. As shown in Table.~\ref{tab:crossdomain}, all methods are only trained on synthetic dataset Scene Flow and evaluated on four real datasets, ETH3D, Middlebury, and KITTI2012/2015 trainsets, with fixed parameters. Our method still achieves the best performance on all four datasets, also surpassing the robust methods DSMNet \cite{zhang2020domain} and CFNet \cite{shen2021cfnet}. Visual comparisons on ETH3D and KITTI2015 trainsets are shown in Fig.~\ref{fig:visual-comparision} and Fig.~\ref{fig:visual-comparision2} respectively.

\begin{table}[ht]
  \centering
  \small
    \caption{Cross-domain generalization evaluation for real-time methods. All methods are only trained on SceneFlow.}
  % \vspace{-1em}
  \setlength{\tabcolsep}{1.5pt}
  \begin{tabular}{l|ccccc}
    \toprule
    \multirow{2}{*}[-2pt]{Method} & {Middlebury} & {KITTI2012} & {KITTI2015}& {ETH3D}\\ 
    \addlinespace[-12pt] \\
    % \cline{2-5}
    \addlinespace[-12pt] \\ 
    & bad 2.0  &  D1-all &  D1-all & bad 1.0 \\
    \midrule
    AANet \cite{xu2020aanet} & 43.8 & 11.3 & 12.6 & 11.4 \\
    AANet+ \cite{xu2020aanet} & 39.4 & 8.8 & 9.0 & 13.1 \\
    BGNet \cite{xu2021bilateral} & 30.4  & 6.2 & 6.6 & 10.1 \\
    HITNet \cite{tankovich2021hitnet} & 28.9  & \textbf{5.9} & \textbf{6.5} & 10.6 \\
    Lite-CREStereo++ & \textbf{27.5} & 6.0 & 7.0 & \textbf{9.9} \\

    \bottomrule
  \end{tabular}
  \label{tab:Efficiency generalization}
% \vspace{-0.8em}
\end{table}

\subsection{Efficiency Evaluation}
Since more iteration numbers lead to increased time costs, we analyze the relationship between iteration numbers and performance. Fig.~\ref{fig:efficiency} (a) shows the experiment conducted with different recurrent methods \cite{li2022practical, lipson2021raft} under similar time costs at a certain iteration. The performance of our method with 6 iterations can outperform other 15 iterations methods, which do not require a large iteration time.

As shown in Table.~\ref{tab:Efficiency evaluation}, we conduct experiments for the proposed lite version method (Lite-CREStereo++) on KITTI2012 and KITTI 2015 online benchmarks. Under similar inference speed to real-time methods, Lite-CREStereo++ achieves SOTA results among all published real-time methods on KITTI2012 benchmark. Meanwhile, it outperforms most published methods on KITTI2015 benchmark at the time of writing.
We also note that most existing methods have $4\times$ more parameters than ours, and our method performs much better than these methods, as shown in Fig.~\ref{fig:efficiency} (b) and (c). We also conduct cross-domain generalization evaluation for existing real-time methods. From Table.~\ref{tab:Efficiency generalization} we can see our method still keeps a high robustness ability, outperforming other methods.

\section{Conclusion}
In this paper, we show that a content-aware warping module based on uncertainty estimation improves the performance of stereo matching, especially on the aspect of robustness. Combined with cascaded architecture and recurrent mechanism, we propose CREStereo++ to recover disparity for robust stereo matching. Moreover, we design a lightweight model with real-time performance. Experimental results show that our approach performs well on various datasets, and has generic applicability. The future direction would be extending our method to other warping-based cost volume tasks, such as multi-view stereo and optical flow.

%%%%%%%%% REFERENCES
{\small
\bibliographystyle{ieee_fullname}
\bibliography{egbib}
}

\end{document}